\documentclass[letterpaper,10pt,conference]{ieeeconf}
\IEEEoverridecommandlockouts
\usepackage{graphicx}
\usepackage{amsmath,amssymb,amsfonts}
\usepackage{booktabs}
\usepackage{multirow}
\usepackage{makecell}
\usepackage{cite}
\usepackage{xcolor}
\usepackage{url}
\usepackage{balance}
\usepackage{dblfloatfix}
\usepackage{algorithm}
\usepackage{algorithmic}

\newcommand{\method}{GRAFT}

\newcommand{\R}{\mathbb{R}}

\title{GRAFT: Grounded and Efficient Online Reinforcement Adaptation for Fine-Grained Robot Manipulation}

\author{
  Yibo Qiu\textsuperscript{$\dagger$}, 
  Haoliang Ye\textsuperscript{$\dagger$}, 
  Shu'ang Sun,
  Zan Huang,
  Ronald X Xu,
  Mingzhai Sun\textsuperscript{$\ast$}
  \thanks{$^\dagger$Equal contribution. $^\ast$Corresponding authors.}
  \thanks{All authors are with the Suzhou Institute for Advanced Research, University of Science and Technology of China, Suzhou, Jiangsu, China, and also with the School of Biomedical Engineering, Division of Life Sciences and Medicine, University of Science and Technology of China, Hefei, Anhui, China.}
  \thanks{Emails: alexandreqiu@mail.ustc.edu.cn; yehubert233@mail.ustc.edu.cn; sunshuang@mail.ustc.edu.cn; huangzan@mail.ustc.edu.cn; xux@ustc.edu.cn; mingzhai@ustc.edu.cn.}
  \thanks{This work was supported by the Gusu Leading Talent Entrepreneurship and Innovation Program (Grant No. ZL2024349).}
}

\begin{document}
\maketitle
\thispagestyle{empty}
\pagestyle{empty}

\begin{abstract}
Pretrained vision-language-action (VLA) policies provide strong priors for robot manipulation, yet adapting them online to fine-grained biomedical tasks remains challenging. Task success often hinges on subtle, view-dependent visual cues, while task-level rewards provide little guidance about which regions matter, making it difficult to learn task-relevant visual grounding from limited real-robot interaction. Online adaptation is further constrained by the computational cost of VLA inference and replay-based updates. We introduce \textbf{GRAFT} (\emph{Grounded Reinforcement Adaptation for Fast Task Learning}), a framework for efficient online VLA adaptation through grounded perception. GRAFT uses region-level supervision to learn view-specific visual anchors that focus perception on task-relevant local cues without requiring region proposals at deployment. It further combines single-step action generation with cached visual-language prefix reuse to accelerate online learning. Across four biomedical manipulation tasks, GRAFT improves success rates by 32.5 percentage points under matched adaptation budgets, while reducing the computational overhead of online policy updates.
\end{abstract}

\section{INTRODUCTION}

Fine-grained laboratory manipulation often depends on precise local contact geometry rather than scene-level recognition alone. Tasks such as attaching a pipette tip, opening a Petri dish lid, transferring liquid, and loading a centrifuge tube require the robot to identify and interact with small task-relevant regions across multiple camera views. Pretrained vision-language-action (VLA) policies provide strong semantic and visuomotor priors for such tasks \cite{black2024pi0,kim2024openvla,wen2025tinyvla}. However, adapting these priors online remains challenging: demonstrations and scalar task-level rewards provide little explicit spatial guidance about which local visual cues are responsible for success or failure. The policy must therefore acquire task-specific fine-grained grounding from limited real-robot interaction, while repeated VLA inference and replay-based optimization impose substantial computational cost.

Recent work has improved spatial reasoning in VLA policies through geometric representations, explicit visual grounding, and auxiliary spatial supervision \cite{li2026pointvla}. Complementary approaches improve VLA adaptation and efficiency through compact architectures, parallel action generation, and reduced computation \cite{wen2025tinyvla,yu2026dmsvla}. These directions address important but largely distinct parts of the problem. For real-robot online adaptation, spatial supervision must rapidly shape task-relevant local representations from limited interaction without becoming an additional dependency at deployment, while repeated acting and learning must remain inexpensive enough to exploit experience within a constrained wall-clock budget. We therefore ask: can region-level supervision teach a VLA which local visual cues matter during adaptation, while allowing the learned representation to remain proposal-free at deployment and efficient to optimize online?

\begin{figure*}[!t]
  \centering
  \includegraphics[width=\textwidth]{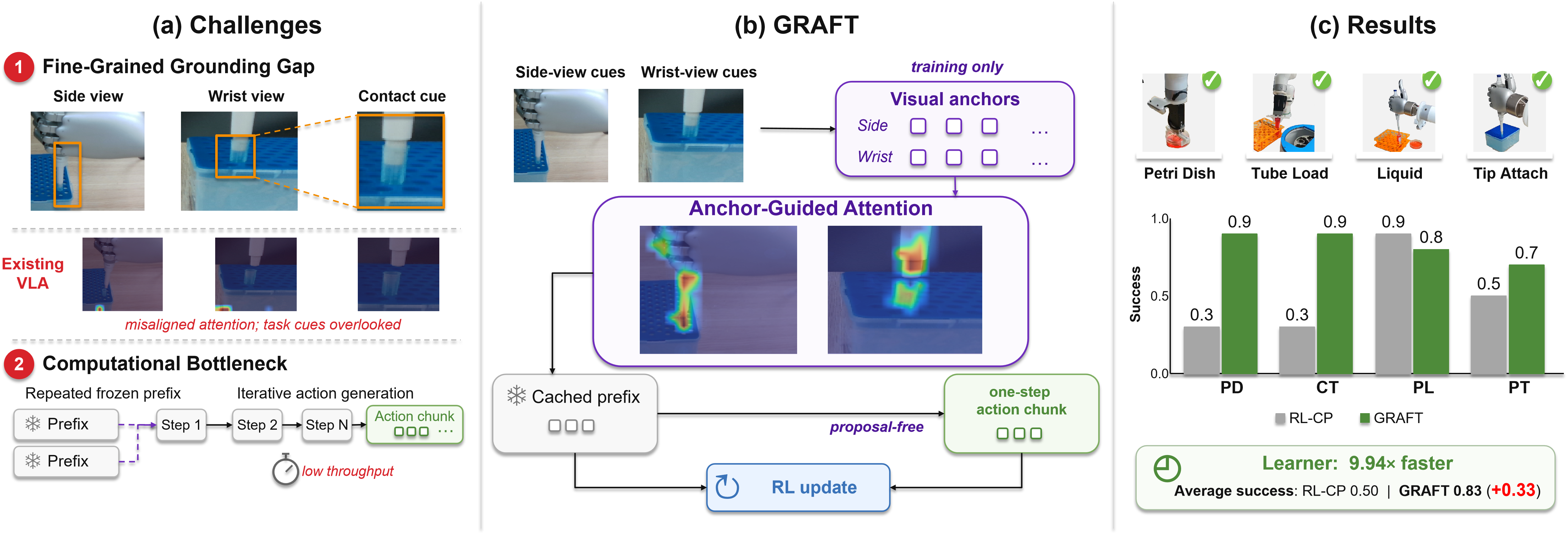}
  \caption{Motivation and overview of GRAFT.
  GRAFT grounds task-relevant visual cues with view-specific anchors and accelerates online adaptation through cached context reuse and single-step action generation.}
  \label{fig:motivation}
\end{figure*}

As illustrated in Figure~\ref{fig:motivation}, we introduce \method{}, or \emph{Grounded Reinforcement Adaptation for Fast Task Learning}, a framework for efficient online VLA adaptation through grounded perception. GRAFT learns view-specific visual anchors that retain task-relevant local information from each camera view. During training, candidate regions provide auxiliary spatial supervision for these anchors. Rather than collapsing multiple candidate regions into a single union target, GRAFT uses identity-free multi-region supervision that preserves distinct plausible regions without requiring a fixed correspondence between anchor identities and region identities. This encourages the visual representation to capture complementary task-relevant local cues across views. Importantly, region supervision is used only during adaptation; at deployment, the policy requires neither region masks nor external proposal generators.

GRAFT also reduces the computational cost of repeatedly acting and learning. On the learner side, the visual-language prefix remains frozen, allowing its key--value (KV) states to be cached and reused across repeated replay updates. The learner therefore recomputes only the trainable anchor-conditioned action pathway, substantially reducing the cost of online optimization. Complementarily, the actor predicts a complete action chunk in a single forward pass for low-latency control. Demonstration and online replay data are jointly optimized using behavior cloning and Q-maximization, allowing the policy to retain demonstrated behavior while improving from online task feedback.

Together, these components target a practical objective for real-robot learning: maximizing task improvement within a fixed wall-clock adaptation budget. We evaluate GRAFT on four fine-grained biomedical manipulation tasks under matched 45-minute adaptation budgets. The experiments measure task success over online adaptation, isolate the effect of multi-region grounding supervision, quantify learner-update efficiency, and visualize the task-relevant spatial structure captured by the learned anchors.

The contributions of this work are:
\begin{itemize}
    \item a multi-view visual anchor mechanism with identity-free multi-region supervision that learns task-relevant local representations during online adaptation while requiring no region proposals or masks at deployment;
    \item an efficient online actor--learner design that reuses frozen-prefix KV states across replay updates and combines this learner-side optimization with single-pass action-chunk prediction for low-latency control;
    \item a matched-budget real-robot evaluation across four fine-grained biomedical manipulation tasks, demonstrating improved final-stage task success together with substantially higher learner-update throughput.
\end{itemize}

\section{RELATED WORK}

\subsection{Robot Learning for Laboratory Automation}

Laboratory manipulation presents several challenges that are less common in standard tabletop settings, including small interaction regions, transparent or reflective objects, specialized instruments, and tightly constrained contact sequences. Recent work has developed simulation environments, benchmarks, and data-generation tools for laboratory robotics, covering chemistry-oriented simulation, biologically grounded benchmarks, editable wet-lab assets, demonstration augmentation, and protocol-conditioned data generation \cite{li2024chemistry3d,lan2025autobio,liu2026pipette,ren2026labvla}. Other systems focus on autonomous laboratory workflows by combining robotic manipulation with protocol reasoning, visual verification, and multimodal feedback \cite{angers2025roboculture,qiu2025biomars,du2026bioprovla}. Reinforcement learning has also been used to improve difficult laboratory skills through structured rewards and online fine-tuning \cite{qiu2026keyframe}. These works mainly address laboratory infrastructure, workflow autonomy, or reward design. Our focus is complementary: we study how a pretrained visuomotor policy can learn which local visual cues matter from limited real-robot interaction.

\subsection{Grounded and Multi-View Perception for VLA Manipulation}

Fine-grained manipulation requires spatial information that may be lost in high-level visual representations. Prior work improves spatial reasoning in VLA policies using 3D-aware representations, point-cloud features, geometry-aware feature transfer, visual traces, and adaptive viewpoint or resolution selection \cite{qu2025spatialvla,li2026pointvla,zhang2025falcon,zheng2024tracevla,liu2026activevla}. Multi-view approaches further combine observations from different cameras through explicit geometric reasoning or cross-view aggregation \cite{fan2026peafowl}.

Another line of work uses explicit grounding signals to make visual features more useful for manipulation. Some methods provide target masks, grounded regions, or high-resolution object crops directly to the policy at execution time \cite{huang2025roboground,yang2026hivla}. Others use gaze, geometric alignment, affordance priors, or auxiliary grounding objectives during training \cite{song2025reconvla,li2025spatialforcing,jia2026guidedvla,kong2026affordvla,zhu2026deltavla}. GRAFT is most closely related to the latter group, but focuses on \emph{online adaptation from multiple robot cameras}. Instead of merging all candidate regions into a single union mask, GRAFT keeps multiple plausible regions and trains lightweight, view-specific anchors with identity-free supervision. The region proposals are used only for supervision and are not needed when the policy is deployed.

\begin{figure*}[!t]
  \centering
  \includegraphics[width=\textwidth]{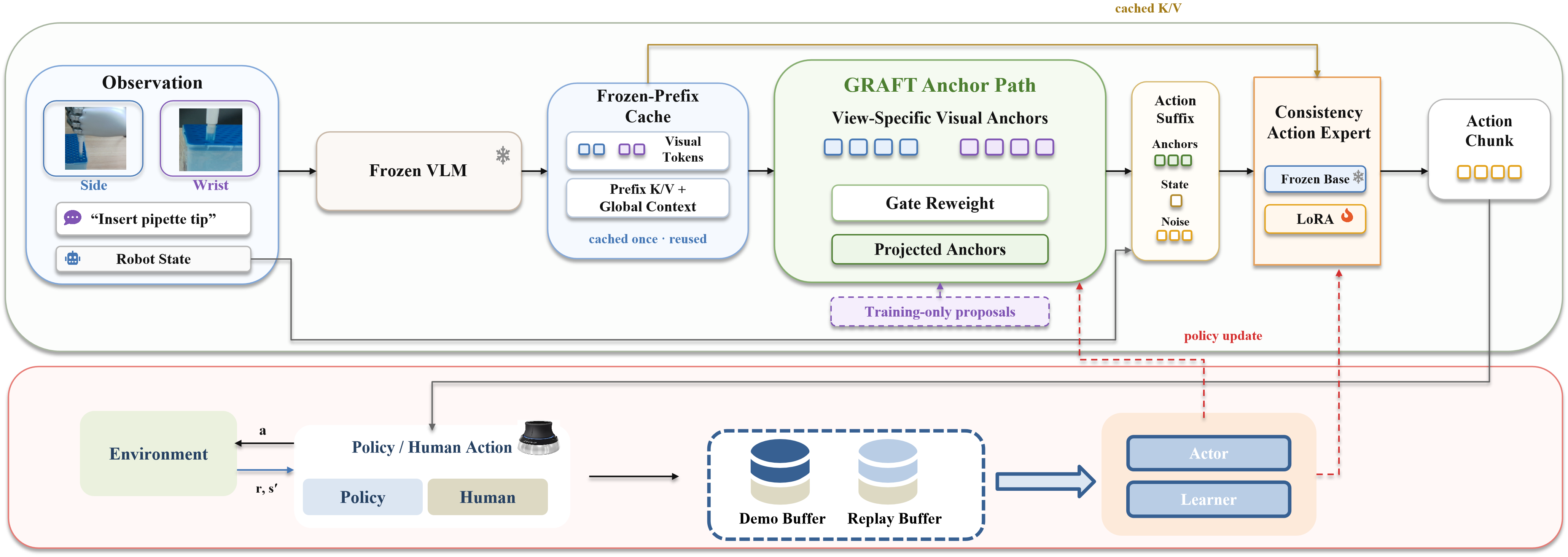}
  \caption{GRAFT architecture and online adaptation pipeline.
  A frozen VLM enables reusable prefix states, visual anchors ground task-relevant cues using training-only proposal supervision, and a consistency action expert generates action chunks in an asynchronous actor--learner loop.}
  \label{fig:method-overview}
\end{figure*}

\subsection{Real-World Reinforcement Adaptation and Efficient VLA Learning}

Real-world reinforcement learning can improve manipulation policies beyond the behaviors covered by demonstrations. Human-in-the-loop off-policy RL has shown that demonstrations and interventions can support efficient learning of precise physical skills \cite{luo2024hilserl}. More recent work adapts pretrained VLA policies using combinations of supervised and reinforcement learning, offline-to-online fine-tuning, chunk-level RL, and improved value estimation \cite{guo2025irevla,chen2025conrft,huang2025corft,zhang2026force,lu2025vlarl,pan2026sop}. Force-aware online post-training additionally uses reactive force feedback and human-corrected rollouts \cite{wang2026lift}. GRAFT builds on this broader line of demonstration-initialized adaptation rather than proposing a new RL objective. Our focus is on making task-relevant visual representations easier to acquire during online learning.

Efficiency is also important because real-robot adaptation is often limited by wall-clock time. Existing methods speed up VLA execution through parallel action decoding and action chunking \cite{kim2025openvlaoft}, while others avoid repeated visual computation by caching or reusing representations across observations and control steps \cite{xu2025vlacache,he2026lifelongvla}. GRAFT exploits a different form of reuse on the learner side. Because the visual-language prefix is frozen during adaptation, its cached states can be reused when replay samples are revisited, reducing repeated computation during learner updates.

\section{METHOD}

Figure~\ref{fig:method-overview} illustrates GRAFT, which combines single-step action-chunk generation and frozen-prefix replay reuse with training-time supervision of view-specialized visual anchors. Region proposals guide spatial grounding only during training and are never queried by the deployed policy.

\subsection{Problem setup and deployment boundary}

At time $t$, the policy receives $s_t=(I_t^s,I_t^w,l,q_t)$, comprising side- and wrist-view RGB observations, language instruction $l$, and adapter state $q_t\in\R^8$. It predicts an action chunk $a_{t:t+H-1}\in\R^{7H}$ with $H=10$, where each low-level action contains translation, rotation, and gripper commands. A frozen VLM prefix maps the inputs to view tokens $X_t^v\in\R^{P_v\times d}$, pooled context $c_t\in\R^d$, and prefix key--value states for $v\in\{s,w\}$.

Training observations may include patch-aligned masks $\mathcal{M}_t^v=\{M_{t,k}^v\}_{k\in\mathcal{V}_v}$ generated offline. Each view supervises only its own anchors; neither masks nor proposal generators are policy inputs.

\subsection{Efficient replay adaptation}

For replayed observations, GRAFT caches the frozen prefix states, pooled summaries, and visual tokens. During learner updates, these states are reused, while all trainable anchor, action, and critic pathways are recomputed; live observations always run a fresh prefix forward pass.

GRAFT replaces iterative flow-matching action generation with a single-step consistency policy~\cite{prasad2024consistency,chen2025conrft}. A single forward pass samples $w\sim\mathcal{N}(0,\sigma^2I)$ and generates $a_{t:t+H-1}\sim\pi_\theta(\cdot\mid s_t)$, retaining stochastic exploration without iterative sampling. The head is trained on demonstration and online-replay chunks.

\subsection{View-specialized visual anchors}

GRAFT maintains separate sets of $N_v=8$ anchor templates for the side and wrist views while sharing the context and cross-attention projections. For $v\in\{s,w\}$,
\begin{align}
\widetilde e_i^v &= e_i^v+\operatorname{NormScale}(W_c c_t), \nonumber\\
A_{i,p}^v &= \operatorname{softmax}_{p}\!\left(
\frac{
\left\langle
\operatorname{norm}(W_q\widetilde e_i^v),
\operatorname{norm}(W_kX_{t,p}^v)
\right\rangle
}{\tau_a}
\right), \nonumber\\
z_i^v &= W_o\sum_{p=1}^{P_v}A_{i,p}^vW_vX_{t,p}^v+b_{\mathrm{view}}^v .
\label{eq:anchor_attention}
\end{align}
Here, $A_i^v$ denotes raw attention and $z_i^v$ the observation-dependent visual anchor. Each anchor reads only its assigned view and has no fixed proposal identity.

Global context reweights anchors within each view:
\begin{equation}
\begin{aligned}
g_i^v
&=
\operatorname{softmax}_{i\in v}\!\left(
\frac{
W_2\operatorname{swish}(W_1[z_i^v;c_t])
}{T_g}
\right),\\
\bar z_i^v
&=
\left[(1-\alpha_g)+\alpha_gN_vg_i^v\right]z_i^v.
\end{aligned}
\label{eq:anchor_reweighting}
\end{equation}
A centered-$\tanh$ adapter with scale $\beta_g$ is applied before suffix fusion.

\subsection{Training-only multi-region guidance}

\begin{figure*}[!t]
  \centering
  \includegraphics[width=\linewidth]{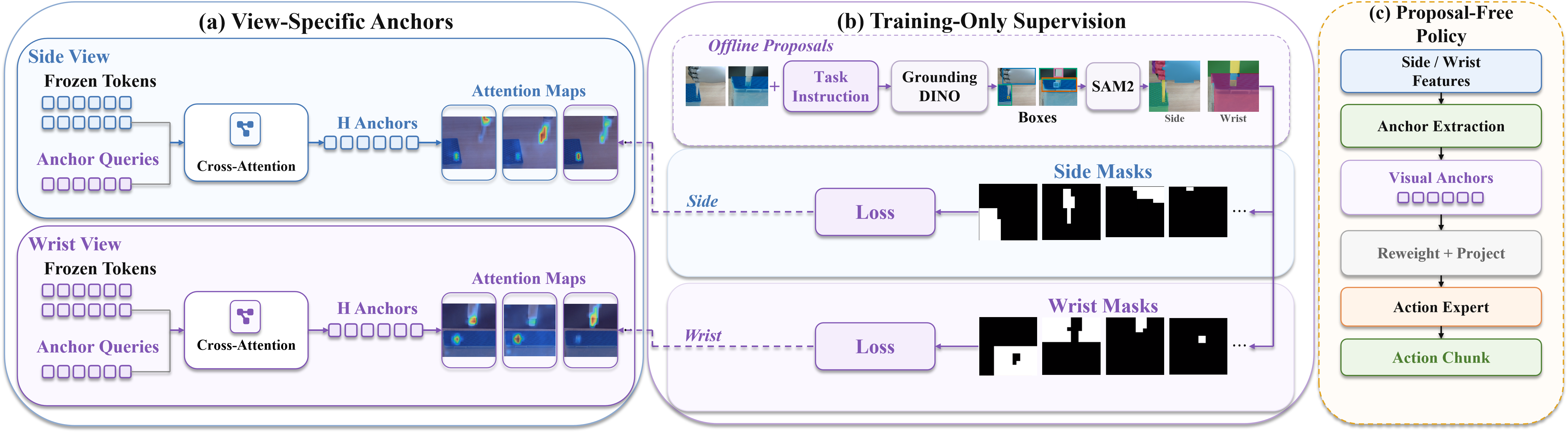}
  \caption{Training-only learning of view-specific visual anchors. Side- and wrist-view anchors read their corresponding frozen visual features. Multiple offline proposal masks guide raw anchor attention independently in each view; the proposal branch is removed at deployment.}
  \label{fig:anchor-supervision}
\end{figure*}

As Fig.~\ref{fig:anchor-supervision} shows, proposal masks supervise raw anchor attention without persistent anchor--proposal assignments. For anchor $i$ and valid proposal $k$ in view $v$,
\begin{equation}
C_{i,k}^v =
1-
\frac{
2\langle A_i^v,M_k^v\rangle+\epsilon
}{
\max\!\left(
\lVert A_i^v\rVert_1+\lVert M_k^v\rVert_1,
\epsilon
\right)
}.
\label{eq:dice_cost}
\end{equation}
Let $\operatorname{smin}_{\tau}\{c_j\}=\sum_j[\operatorname{softmax}(-\mathbf{c}/\tau)]_jc_j$. FreeDice matches proposals to anchors and anchors to valid proposals:
\begin{equation}
\begin{aligned}
\mathcal{L}_{\mathrm{FD}}^v ={}&
\frac{1}{|\mathcal{V}_v|}
\sum_{k\in\mathcal{V}_v}
\operatorname{smin}_{\tau_D}
\{C_{i,k}^v\}_{i=1}^{N_v}\\
&+
\frac{\lambda_{\mathrm{anc}}}{N_v}
\sum_{i=1}^{N_v}
\operatorname{smin}_{\tau_D}
\{C_{i,k}^v\}_{k\in\mathcal{V}_v}.
\end{aligned}
\label{eq:freedice}
\end{equation}
The two terms respectively encourage proposal coverage and prevent anchors from ignoring all valid regions; matching is rebuilt for every frame and view.

With $\widetilde A_i^v$ denoting attention normalized over the valid-proposal union, the redundancy penalty and grounding loss are
\begin{equation}
\begin{aligned}
\mathcal{L}_{\mathrm{dup}}
&=
\underset{v,(i,j)\in\mathcal{P}_v}{\operatorname{mean}}
\left[
\langle\widetilde A_i^v,\widetilde A_j^v\rangle
-m_{\mathrm{dup}}
\right]_+,\\
\mathcal{L}_{\mathrm{ground}}
&=
\alpha
(\mathcal{L}_{\mathrm{FD}}^s+\mathcal{L}_{\mathrm{FD}}^w)
+\beta\mathcal{L}_{\mathrm{dup}}.
\end{aligned}
\label{eq:grounding_loss}
\end{equation}
We use one configuration across tasks: $(\tau_a,T_g,\alpha_g,\beta_g)=(0.5,1.5,0.1,0.04)$ and $(\lambda_{\mathrm{anc}},m_{\mathrm{dup}},\alpha,\beta)=(0.25,0.90,0.4,0.5)$, with fixed $\tau_D$. Views without valid proposals contribute zero.

\subsection{Online adaptation and proposal-free inference}

GRAFT runs asynchronous actor and learner processes. The actor uploads interactions and corrections at episode boundaries. Once online replay contains 100 valid windows, the learner samples equally from online and demonstration data. At replanning time, the actor samples $a_{t:t+H-1}\sim\pi_\theta(\cdot\mid s_t)$ and executes its first $e$ actions. Replay windows may cross replanning blocks and include human corrections. Each transition is written as $(s,a,r,s',m)\sim\mathcal{D}$, where $a\in\R^{7H}$ is the executed window, $s'=s_{t+H}$, $r$ is its return, $m$ is the bootstrap mask, and $\mathcal{D}$ denotes balanced sampling from both buffers. All valid windows enter online replay, and intervention windows also enter the demonstration buffer.

\noindent\textit{1) Critic update:} With twin critics $Q_{\phi}$, the target action and Bellman target are
\begin{equation}
a'\sim\pi_\theta(\cdot\mid s'), \qquad
y=r+\gamma m\min_i Q_{\bar\phi_i}(s',a').
\label{eq:critic_target}
\end{equation}
For $\gamma=0.98$, each critic minimizes
\begin{equation}
\mathcal{L}_{\mathrm{critic}}
=
\mathbb{E}_{(s,a,r,s',m)\sim\mathcal{D}}
\left[
\left(Q_\phi(s,a)-y\right)^2
\right].
\label{eq:critic_objective}
\end{equation}

\noindent\textit{2) Actor update:} The actor combines behavior cloning and value maximization:
\begin{equation}
\begin{aligned}
\mathcal{L}_{\mathrm{actor}}
&=
\lambda_{\mathrm{bc}}\mathcal{L}_{\mathrm{BC}}
+\lambda_q\mathcal{L}_Q,\\
\mathcal{L}_{\mathrm{BC}}
&=
\mathbb{E}_{(s,a)\sim\mathcal{D}}
\left[
\left\lVert a-\pi_\theta(s)\right\rVert_2^2
\right],\\
\mathcal{L}_Q
&=
-\mathbb{E}_{s\sim\mathcal{D}}
\left[
Q_\phi\!\left(s,\pi_\theta(s)\right)
\right].
\end{aligned}
\label{eq:actor_objective}
\end{equation}
We use $\lambda_{\mathrm{bc}}=0.5$ and $\lambda_q=1$. Each iteration performs two critic updates and one actor update; every ten iterations, proposal-valid demonstration windows provide an actor-only $\mathcal{L}_{\mathrm{ground}}$ update. Algorithm~\ref{alg:graft_online_adaptation} summarizes the procedure.

\begin{algorithm}[t]
\footnotesize
\caption{GRAFT Online Policy Adaptation}
\label{alg:graft_online_adaptation}
\begin{algorithmic}[1]
\REQUIRE
$\mathcal{D}_{\mathrm{demo}},\mathcal{D}_{\mathrm{online}}$: demonstration and replay buffers; \\
$\pi_0$: pretrained VLA; $Q_{\phi_1},Q_{\phi_2}$: twin critics; \\
$\mathcal{C}$: Prefix-KV cache; $H,e$: action horizon and execution length; \\
$G,P$: grounding and policy-publication periods.

\STATE Initialize $\pi_\theta\leftarrow\pi_0$, $\bar\phi_i\leftarrow\phi_i$ $(i=1,2)$, and $k\leftarrow0$.
\STATE Use the following logical Actor--Learner schedule.
\FOR{each episode}
    \STATE Reset the robot and load the latest $\pi_\theta$; $\mathcal{Q}\leftarrow\varnothing$.
    \WHILE{the episode is not terminated}
        \IF{$\mathcal{Q}=\varnothing$}
            \STATE Observe $s_t$ and sample $a_{t:t+H-1}\sim\pi_\theta(\cdot\mid s_t)$.
            \STATE Enqueue its first $e$ actions in $\mathcal{Q}$.
        \ENDIF
        \STATE $a_t\leftarrow\operatorname{Pop}(\mathcal{Q})$, $h_t\leftarrow0$.
        \IF{human intervention occurs}
            \STATE $a_t\leftarrow a_t^*$, $h_t\leftarrow1$.
        \ENDIF
        \STATE Execute $a_t$; record $\xi_t=(s_t,a_t,r_t,s_{t+1},d_t,h_t)$.
    \ENDWHILE
    \STATE Form valid windows $\omega_j=(s_j,a_{j:j+H-1}^{\mathrm{exec}},R_j,s_{j+H},m_j)$.
    \STATE Add all $\omega_j$ to $\mathcal{D}_{\mathrm{online}}$.
    \STATE Add intervention windows to $\mathcal{D}_{\mathrm{demo}}$ and upload the episode.
    \IF{$|\mathcal{D}_{\mathrm{online}}|\geq100$}
        \STATE Sample equal-size $\mathcal{B}_{\mathrm{on}}\sim\mathcal{D}_{\mathrm{online}}$ and $\mathcal{B}_{\mathrm{demo}}\sim\mathcal{D}_{\mathrm{demo}}$.
        \STATE $\mathcal{B}\leftarrow\mathcal{B}_{\mathrm{on}}\cup\mathcal{B}_{\mathrm{demo}}$; reuse frozen-prefix states in $\mathcal{C}$.
        \STATE Recompute trainable pathways; update each critic twice with $\mathcal{L}_{\mathrm{critic}}$.
        \STATE Polyak-update targets; update $\pi_\theta$ with $\mathcal{L}_{\mathrm{actor}}$; $k\leftarrow k+1$.
        \IF{$k\bmod G=0$ and proposal-valid windows are available}
            \STATE $\mathcal{B}_{\mathrm{prop}}\sim\mathcal{D}_{\mathrm{demo}}^{\mathrm{prop}}$; update the actor with $\mathcal{L}_{\mathrm{ground}}$.
        \ENDIF
        \IF{$k\bmod P=0$}
            \STATE Publish $\pi_\theta$ to the Actor.
        \ENDIF
    \ENDIF
\ENDFOR
\end{algorithmic}
\end{algorithm}

At deployment, policy inference uses only observations, language, adapter state, learned anchors, and the action head. Proposal generators, masks, grounding losses, and critics are absent.

\section{EXPERIMENTS}

We structure our evaluation around four questions: \textbf{Q1:} Does identity-free multi-region anchor supervision improve fine-grained manipulation over proposal-union supervision? \textbf{Q2:} Does GRAFT achieve stronger online adaptation under a fixed interaction budget? \textbf{Q3:} How much do single-step action generation and prefix reuse reduce learner-update cost? \textbf{Q4:} Does training-time multi-region supervision improve task-relevant spatial grounding while preserving proposal-free deployment? We address these questions through controlled real-robot comparisons, online adaptation trajectories, learner-throughput measurements, and grounding visualizations.

\subsection{Real-robot tasks and protocol}

We evaluate four fine-grained manipulation tasks on a JAKA arm: Petri Dish De-lidding, Centrifuge Tube Loading, Precision Liquid Transfer, and Pipette Tip Attachment. The policy receives side-view and wrist-view RGB observations together with proprioception. All policies predict an action horizon of $H=10$. Pipette Tip Attachment replans after every three executed actions, whereas the other tasks execute the full 10-action chunk before replanning.

For each task, we collect 10 teleoperated demonstrations and allocate 45 minutes of online adaptation to each method. Episodes start from a fixed robot configuration, while task objects are randomly perturbed within $\pm 30\,\mathrm{mm}$ of their nominal planar positions. Success is determined by predefined task-specific criteria. Human intervention is allowed only when autonomous recovery is unlikely; intervened episodes are logged separately and are not counted as autonomous successes.

\begin{figure*}[!t]
  \centering
  \includegraphics[width=\textwidth]{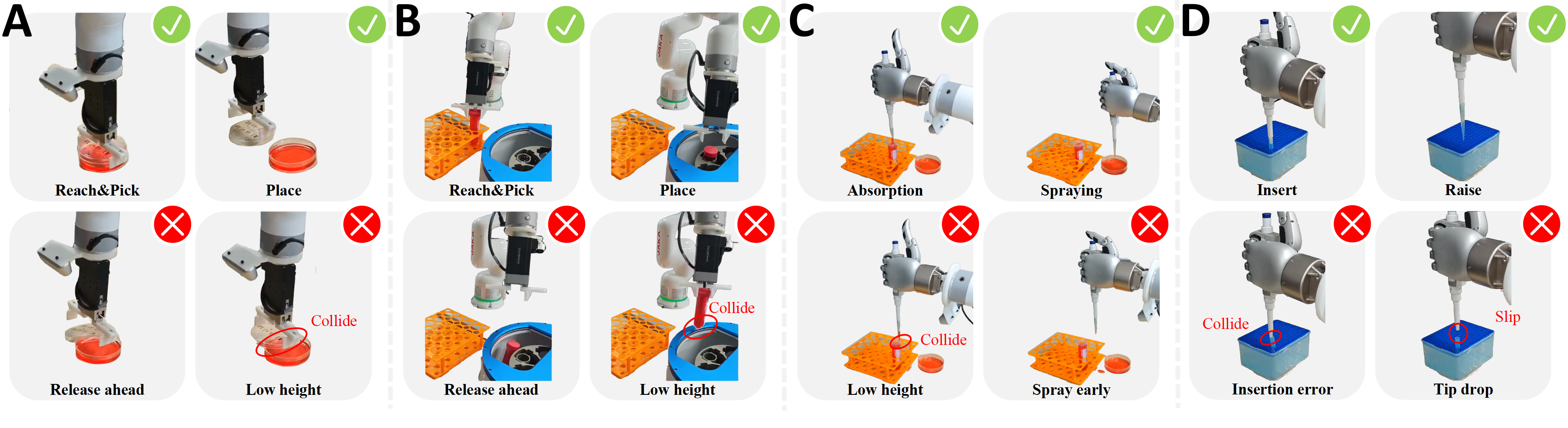}
  \caption{Qualitative real-robot executions and representative failure modes in the four evaluation tasks. (A) Petri Dish De-lidding, (B) Centrifuge Tube Loading, (C) Precision Liquid Transfer, and (D) Pipette Tip Attachment. The upper row in each panel shows key stages of a successful execution, and the lower row highlights representative failure modes.}
  \label{fig:task-success-failure}
\end{figure*}

Figure~\ref{fig:task-success-failure} shows representative success and failure cases. Common failures include inaccurate alignment, premature release, low-height contact, collision, and insertion errors, highlighting the spatial precision required by these tasks.

\subsection{Baselines and controlled ablations}

All methods use the same demonstrations, interaction budget, action horizon, robot hardware, and evaluation protocol. \textbf{RL-FM} uses the native multi-step flow-matching action generator, whereas \textbf{RL-CP} replaces it with a single-step consistency head. Their comparison isolates the effect of action generation on online adaptation and learner-update efficiency.

\textbf{GRAFT-Union} introduces view-specialized anchors, context reweighting, overlap regularization, and proposal-union supervision. \textbf{GRAFT} keeps all other components unchanged and replaces only proposal-union supervision with identity-free multi-region FreeDice. The GRAFT-Union versus GRAFT comparison therefore isolates the effect of the grounding supervision objective.

\subsection{Task success and online adaptation}

\begin{table}[t]
\centering
\caption{Online rolling success in the final ten episodes of each available trace.}
\label{tab:online_rolling_success}
\setlength{\tabcolsep}{0.4pt}
\renewcommand{\arraystretch}{1.05}
\tiny
\begin{tabular*}{\columnwidth}{@{\extracolsep{\fill}}lccccc@{}}
\toprule
Policy & PD & CT & PL & PT & Overall \\
\midrule
RL-FM
& 2/10 & 0/10 & 0/10 & 2/10
& 4/40 (10.0\%) \\

RL-CP
& 3/10 & 3/10 & 9/10 & 5/10
& 20/40 (50.0\%) \\

GRAFT-Union
& 7/10 & 2/10 & 8/10 & 6/10
& 23/40 (57.5\%) \\

GRAFT
& 9/10 & 9/10 & 8/10 & 7/10
& 33/40 (82.5\%) \\
\bottomrule
\end{tabular*}
\par\vspace{5pt}
\parbox{\columnwidth}{%
\tiny
PD: Petri Dish De-lidding;
CT: Centrifuge Tube Loading;
PL: Precision Liquid Transfer;
PT: Pipette Tip Attachment.%
}
\end{table}

\begin{figure*}[!b]
  \centering
  \includegraphics[width=\textwidth]{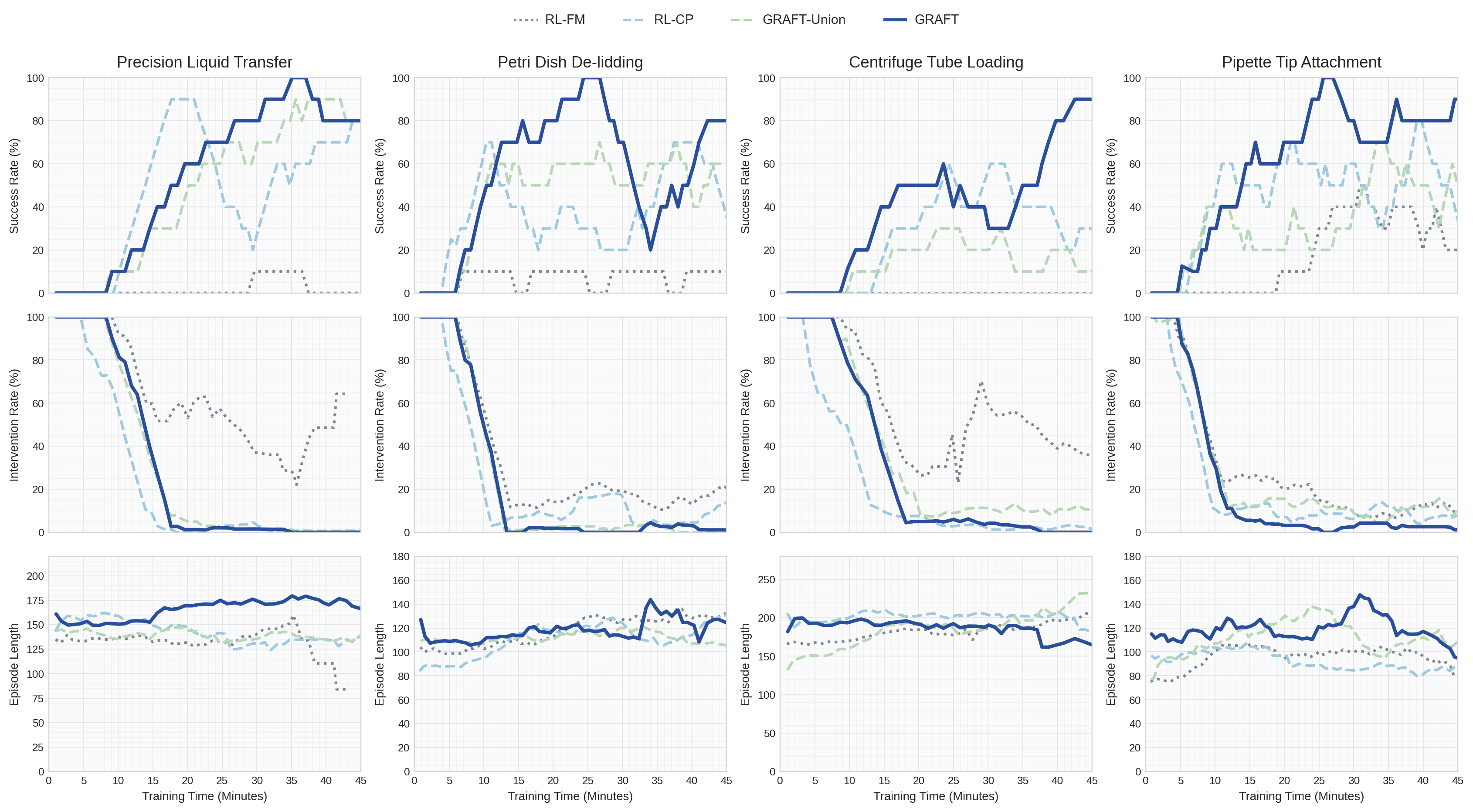}
  \caption{Online adaptation trajectories on the four real-robot tasks. Rows show rolling success rate, intervention rate, and episode length against wall-clock training time for RL-FM, RL-CP, GRAFT-Union, and GRAFT.}
  \label{fig:online-adaptation-curves}
\end{figure*}

Table~\ref{tab:online_rolling_success} reports rolling success over the final 10 episodes of each task, while Figure~\ref{fig:online-adaptation-curves} shows the corresponding online adaptation trajectories and intervention rates. The table summarizes the endpoints of the online trajectories rather than results from a separate frozen-policy evaluation.

Under the same 45-minute interaction budget, GRAFT achieves 33 successes across the 40 final episodes (82.5\%), compared with 23/40 (57.5\%) for GRAFT-Union, an absolute improvement of 25 percentage points. The largest gain occurs on Centrifuge Tube Loading, where success increases from 2/10 to 9/10. GRAFT also improves on two additional tasks, while both variants achieve 8/10 on Precision Liquid Transfer. Since GRAFT and GRAFT-Union differ only in their grounding supervision, this controlled ablation supports the benefit of identity-free multi-region FreeDice over proposal-union supervision for fine-grained manipulation.

The online trajectories show the clearest late-stage separation on Centrifuge Tube Loading. GRAFT also recovers to 90\% success after a mid-training drop on Petri Dish De-lidding and achieves stronger late-stage performance on Pipette Tip Attachment. On Precision Liquid Transfer, however, RL-CP reaches high success earlier and finishes at 9/10, compared with 8/10 for GRAFT. Thus, GRAFT's advantage lies in stronger overall final-stage adaptation under the fixed interaction budget rather than uniformly faster convergence across tasks.

The improvement in success does not rely on increased human assistance. GRAFT approaches near-zero intervention on three tasks and maintains a low intervention rate on Pipette Tip Attachment, indicating that the performance gain primarily comes from improved autonomous behavior.

Across the four tasks, GRAFT attains at least 7/10 successes, showing consistent late-stage performance across manipulation settings with different geometric and contact requirements. The largest gain occurs on Centrifuge Tube Loading, which requires simultaneous localization of the tube, gripper, and loading target. This pattern suggests that separate supervision of multiple local regions is particularly useful when success depends on coordinating several spatial cues.

\subsection{Learner-update efficiency}

We measure learner-update throughput on Centrifuge Tube Loading using 10 demonstrations, 946 replay chunks, and 1,000 learner steps per configuration. Prefix-KV caching reuses frozen-prefix computation for repeated replay observations.

\begin{table}[t]
\centering
\caption{Offline learner-update efficiency on Centrifuge Tube Loading with and without prefix-KV caching, where $K$ is the number of action-generation steps.}
\label{tab:pickup_action_generation_efficiency}
\setlength{\tabcolsep}{3.5pt}
\renewcommand{\arraystretch}{1.05}
\footnotesize
\resizebox{\columnwidth}{!}{%
\begin{tabular}{lcccc}
\toprule
Policy formulation & Prefix KV cache & Total learner (s) & Steady (steps/s) & Steady (s/step) \\
\midrule
RL-FM ($K{=}10$)
& Off & 597.989 & 1.990 & 0.503 \\

RL-FM ($K{=}10$)
& On & 236.405 & 6.300 & 0.159 \\

RL-CP ($K{=}1$)
& Off & 542.047 & 2.210 & 0.452 \\

RL-CP ($K{=}1$)
& On & 119.216 & 21.960 & 0.0455 \\
\bottomrule
\end{tabular}%
}
\end{table}

Without caching, RL-FM and RL-CP achieve 1.99 and 2.21 learner steps/s, respectively. Prefix reuse increases RL-FM throughput to 6.30 steps/s, corresponding to a $3.17\times$ speedup, and increases RL-CP throughput to 21.96 steps/s, corresponding to a $9.94\times$ speedup. With caching enabled, RL-CP is additionally $3.49\times$ faster than RL-FM.

These results show complementary gains from the two efficiency mechanisms: single-step generation reduces action-generation cost, while prefix reuse avoids repeated computation over replayed observation prefixes. Their combination yields the highest learner-update throughput.

\subsection{Grounding diagnostics and deployment}

\begin{figure}[!t]
  \centering
  \includegraphics[width=\columnwidth]{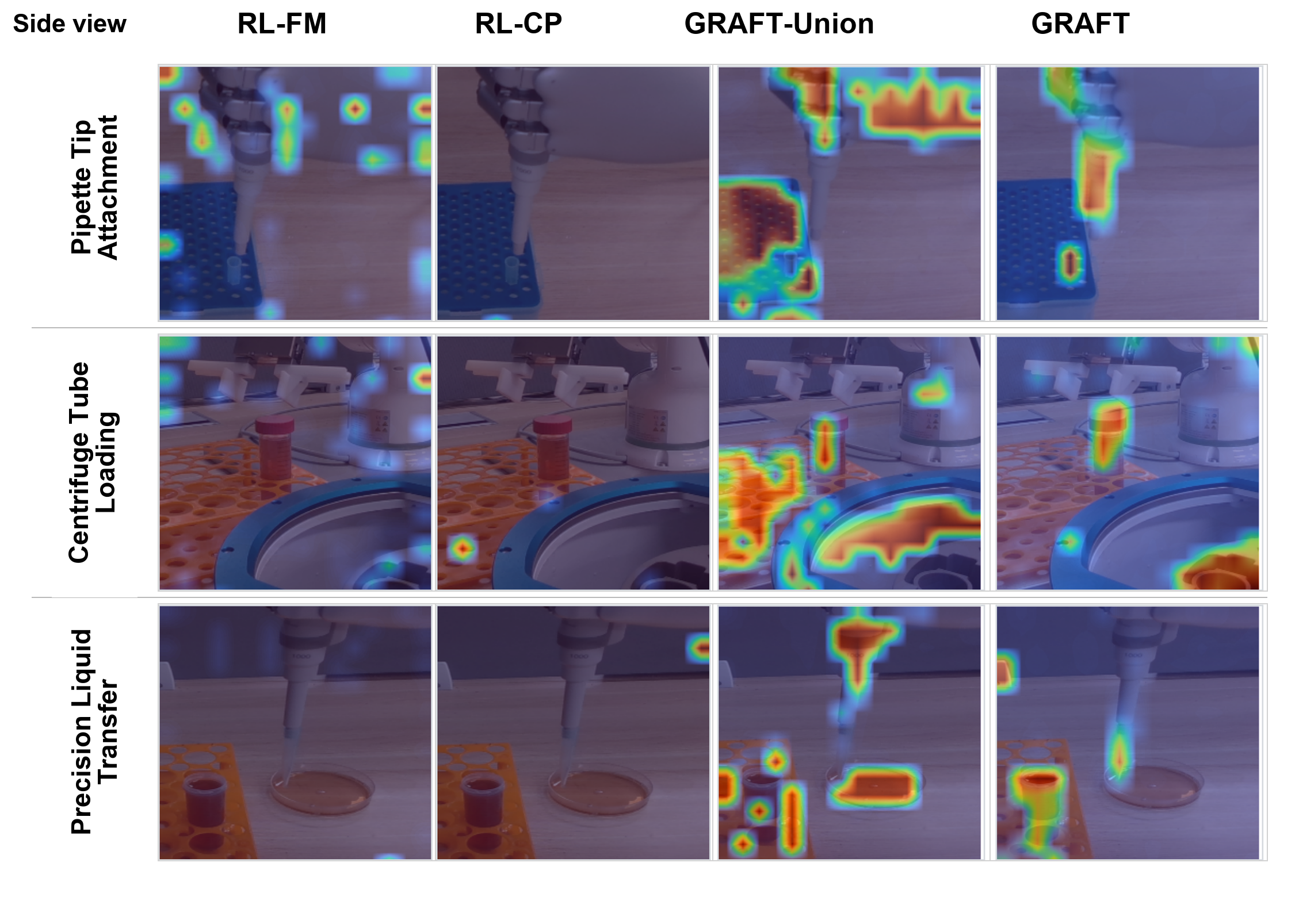}
  \caption{Qualitative comparison of view-specific attention on three representative tasks. The panels show input images, training-only proposal masks, baseline VLA attention, and GRAFT anchor attention for the side and wrist views. Colored indicators denote the visual anchors emphasized in each view.}
  \label{fig:anchor-specialization}
\end{figure}

Figure~\ref{fig:anchor-specialization} visualizes spatial attention on three tasks. For RL-FM and RL-CP, we show final-layer action-token-to-patch attention; for GRAFT-Union and GRAFT, we show raw anchor-to-patch attention. Because these are different attention quantities, cross-family comparisons are qualitative. In contrast, GRAFT-Union and GRAFT visualize the same quantity and differ only in their grounding supervision, providing a controlled comparison of the two objectives.

Compared with GRAFT-Union, GRAFT produces more spatially concentrated responses around manipulation-relevant regions, including pipette tips, centrifuge tubes, and liquid-transfer targets. This qualitative specialization is consistent with the performance gains observed in the controlled comparison above.

The two views exhibit complementary spatial roles: side-view anchors emphasize global approach and object alignment, whereas wrist-view anchors concentrate on local contact and insertion regions, consistent with the intended view-specialized design.

Proposal generators and region masks are used only during training and are removed entirely at policy inference. GRAFT therefore benefits from multi-region grounding supervision without introducing proposal dependence during deployment.





\section{CONCLUSION}

We presented GRAFT, a framework for online adaptation of VLA policies to fine-grained manipulation tasks. GRAFT uses region supervision during training to help view-specific anchors capture local visual cues that are important for the task. These regions are needed only for supervision: at test time, the policy operates directly on the original camera observations without a proposal generator. We also use single-step action generation and prefix-KV caching to make online learning more efficient.

Across four real-robot tasks, GRAFT improves final-stage success by 25 percentage points under the same 45-minute adaptation budget. Prefix caching speeds up learner updates by up to $9.94\times$. The learned anchors also attend to task-relevant regions such as pipette tips, tubes, and liquid-transfer targets, consistent with the improvements observed in our controlled comparison.

One limitation of GRAFT is that the policy is reactive and does not explicitly keep track of previous actions or task progress. Incorporating temporal context may therefore help with longer-horizon tasks in which the relevant visual target changes over time.

\balance
\bibliographystyle{IEEEtran}
\bibliography{root}
\end{document}